\documentclass{ieeetj}
\usepackage{cite}
\usepackage{amsmath,amssymb,amsfonts}
\usepackage{algorithmic}
\usepackage{graphicx,color}
\usepackage{textcomp}
\usepackage{xcolor}
\usepackage{hyperref}
\hypersetup{hidelinks=true}
\usepackage{algorithm,algorithmic}
\usepackage{bm}

\usepackage[small,bf]{caption}
\usepackage{wrapfig}
\usepackage{lipsum}
\usepackage{mathtools}
\usepackage{mwe}
\usepackage{multirow}
\usepackage{tikz}
\usepackage[subrefformat=parens]{subcaption}
\usepackage{setspace}
\usepackage{colortbl}
\usepackage{wrapfig}
\usepackage{amsmath,adjustbox}

\usepackage[utf8]{inputenc}
\usepackage[T1]{fontenc}

\usepackage{amsthm}
\theoremstyle{plain}

\newcommand{\best}[2]{\underline{\textbf{#1}}$\pm$\scriptsize{#2}}

\newcommand{\third}[2]{#1$\pm$\scriptsize{#2}}

\def\BibTeX{{\rm B\kern-.05em{\sc i\kern-.025em b}\kern-.08em
    T\kern-.1667em\lower.7ex\hbox{E}\kern-.125emX}}
\AtBeginDocument{\definecolor{tmlcncolor}{cmyk}{0.93,0.59,0.15,0.02}\definecolor{NavyBlue}{RGB}{0,86,125}}

\def\OJlogo{\vspace{-4pt}$<$Society logo(s) and publication title will appear here.$>$}
\def\seclogo{\vspace{10pt}$<$Society logo(s) and publication title will appear here.$>$}

\def\authorrefmark#1{\ensuremath{^{\textbf{#1}}}}

\begin{document}
\receiveddate{XX Month, XXXX}
\reviseddate{XX Month, XXXX}
\accepteddate{XX Month, XXXX}
\publisheddate{XX Month, XXXX}
\currentdate{XX Month, XXXX}
\doiinfo{XXXX.2022.1234567}

\markboth{}{Author {et al.}}

\title{Representation Synthesis by Probabilistic Many-Valued Logic Operation in Self-Supervised Learning}

\author{
Hiroki Nakamura\authorrefmark{1}, 
Masashi Okada\authorrefmark{1},\\ 
Tadahiro Taniguchi\authorrefmark{1,2}
}
\affil{Panasonic Holdings, Kadoma, Osaka 571-8501 JAPAN}
\affil{Ritsumeikan University, Kyoto, Kyoto 604-8520 JAPAN}
\corresp{Corresponding author: Hiroki Nakamura (email: nakamura.hiroki003@jp.panasonic.com}

\begin{abstract}
In this paper, we propose a new self-supervised learning (SSL) method for representations that enable logic operations.
Representation learning has been applied to various tasks like image generation and retrieval.
The logical controllability of representations is important for these tasks.
Although some methods have been shown to enable the intuitive control of representations using natural languages as the inputs, representation control via logic operations between representations has not been demonstrated.
Some SSL methods using representation synthesis (e.g., elementwise mean and maximum operations) have been proposed, but the operations performed in these methods do not incorporate logic operations.
In this work, we propose a logic-operable self-supervised representation learning method by replacing the existing representation synthesis with the OR operation on the probabilistic extension of many-valued logic.
The representations comprise a set of feature-possession degrees, which are truth values indicating the presence or absence of each feature in the image, and realize the logic operations (e.g., OR and AND).
Our method can generate a representation that has the features of both representations or only those features common to both representations.
Furthermore, the expression of the ambiguous presence of a feature is realized by indicating the feature-possession degree by the probability distribution of truth values of the many-valued logic.
We showed that our method performs competitively in single and multi-label classification tasks compared with prior SSL methods using synthetic representations.
Moreover, experiments on image retrieval using MNIST and PascalVOC showed the representations of our method can be operated by OR and AND operations.
\end{abstract}

\begin{IEEEkeywords}
logic operation, 
operability,
representation learning,
self-supervised learning
\end{IEEEkeywords}


\maketitle

\section{INTRODUCTION}
Remarkable progress has been achieved in representation learning recently. 
Self-supervised learning (SSL) is a method for learning representations without labeled data~\cite{chen2021exploring, grill2020bootstrap, he2020momentum, chen2021empirical, chen2020simple, zbontar2021barlow, tian2021understanding, newell2020useful, komodakis2018unsupervised, noroozi2016unsupervised, Nakamura_2023_ICCV}.
Using models pre-trained by SSL for image classification and object detection achieves high performance with a small amount of labeled data.
Popular SSL methods, e.g., BYOL~\cite{grill2020bootstrap}, SimSiam~\cite{chen2021exploring}, and DINO~\cite{caron2021emerging}, learn a model to maximize the similarity between representations of augmented views of an image.
Owing to these advances in representation learning, various foundation models~\cite{radford2021learning, oquab2023dinov2, brown2020language, pmlr-v139-ramesh21a} have been proposed.
It is a pre-trained model built on large-scale datasets and can be adapted to various tasks and domains.
CLIP~\cite{radford2021learning} achieves zero-shot image classification by learning the common representation space of the two modalities: natural language and images.
DALL-E~\cite{pmlr-v139-ramesh21a} generates images by the diffusion model from a representation of text descriptions.

\begin{figure}[t]
    \centering
    \includegraphics[width=0.96\linewidth]{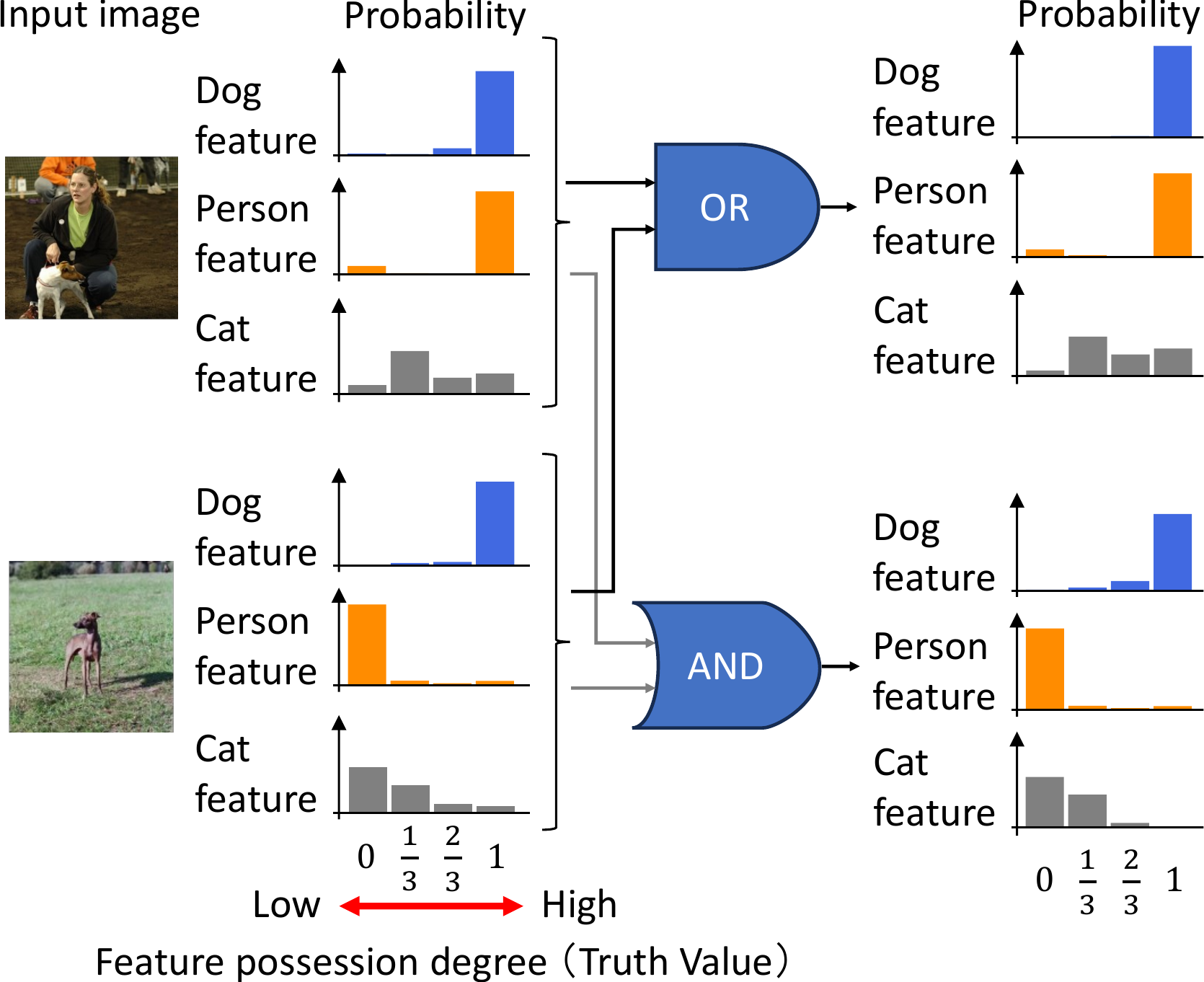}
    \caption{
        \textbf{Overview of the representations and logic operation.}
        The method predicts a representation capable of performing logic operations between representations using many-valued logic. 
        The representation comprises several probability distributions that express the probability of the truth value of the feature-possession degree.
        These examples show three categorical distributions, possibly corresponding to ``Dog'', ``Person'' and ``Cat.''
        Logic operations (e.g., OR, AND) between representations can be performed by presenting feature possession in many-valued logic.
        }
    \label{fig:abstract_method}
\end{figure}
%

%
Control of representations, including the logic operations of representations, is important in various applications such as image generation and retrieval~\cite{dey2019doodle, ramesh2022hierarchical, sueyoshi2023predicated, pmlr-v139-ramesh21a, havaei2021conditional, gonzalez2018image, guo2018dialog, levy2024chatting, ding2020guided, higgins2016beta, chen2016infogan, rassin2024linguistic}. 
Controlling representations using natural language has facilitated intuitive control of representations~\cite{pmlr-v139-ramesh21a, radford2021learning,sueyoshi2023predicated, rassin2024linguistic}.
However, it has been pointed out that image generation methods using natural language often inadequately reflect the logical relationships within the prompt~\cite{sueyoshi2023predicated, rassin2024linguistic, truong-etal-2023-language}. 
Additionally, methods for controlling representations using human feedback and representation disentanglement have been proposed~\cite{ding2020guided, higgins2016beta, gonzalez2018image, havaei2021conditional, levy2024chatting, guo2018dialog}, but few methods explicitly perform representation control compatible with propositional logic operations. 
If operations corresponding to the logical OR and AND operations could be realized, it would be possible to generate representations that combine features of both representations and representations that share only common features. 
We believe that achieving such logic operations between representations will expand the range of representation control and enhance the convenience of various applications utilizing representations.
%

In SSL, several methods use representations generated by operations between representations~\cite{shen2022mix, ren2022simple, lee2021imix, kim2020mixco, zhang2022mix}.
As representation synthesis methods, elementwise mean~\cite{ren2022simple} and elementwise maximum~\cite{zhang2022mix} operations of two representations have been proposed.
The primary objective of these methods is to learn diverse and robust features of images using mixed images~\cite{zhang2018mixup, yun2019cutmix}, and they have not been designed to enhance the operability of representations. 
As a result, these methods have not incorporate the possibility of logic operations. 

In this study, we propose a logic-operable self-supervised representation learning method to enhance the operability of representations.
We realize the control of representations via logic operations by applying the OR operation of the probabilistic extension of the G\"{o}del logic~\cite{godel1932intuitionistischen}, which is one of many-valued logic, to the representation synthesis operation of the conventional SSL methods.
A representation of our method comprises degrees of feature possession, where the feature-possession degree indicates the degree to which each feature is included in the input image. 
Logic operations between representations can be performed by expressing the feature-possession degree in truth value.
In addition,  using the probabilistic extension of many-valued logic, the feature-possession degree can be presented by the categorical distribution of truth values and the expression of the ambiguous presence of features can be achieved.
Fig.~\ref{fig:abstract_method} shows examples of the feature-possession degree and the logic operation of representations.
This method achieves operable representation learning by maximizing the similarity between the representation of a mixture of two images and the representation of each image synthesized by the OR operation.
The contributions of this study can be summarized as follows:
\begin{itemize} 
    \item 
    We proposed a new self-supervised representation learning method that enables logic operation between representations.
    \item 
    We conducted experiments on single and multi-label classification tasks and demonstrated that our method is competitive with existing methods.
    \item 
    We demonstrated that the representations successfully express the degree of feature possession without labels.
    In addition, the image retrieval results showed that the representation of our method is operable by OR and AND operations.
\end{itemize} 
\section{RELATED WORK}
\subsection{SSL and Representation Synthesis}
Previous studies have shown that SSL performs well in many downstream tasks, such as classification, object detection, and semantic segmentation~\cite{chen2020simple, he2020momentum, chen2021exploring, grill2020bootstrap, caron2021emerging}.
SimCLR\cite{chen2020simple} and MoCo~\cite{he2020momentum}, called contrastive learning, learn to maximize the similarity of representation pairs augmented from the same image (\textit{positive pairs}) and to minimize the similarity of representation pairs augmented from different images (\textit{negative pairs}).
Meanwhile, SimSiam~\cite{chen2021exploring}, BYOL \cite{grill2020bootstrap}, and DINO~\cite{caron2021emerging}, called non-contrastive learning, learn using only the positive pairs.

Recently, several methods~\cite{shen2022mix, ren2022simple, lee2021imix, kim2020mixco} that use representations synthesized from multiple representations have been proposed.
Ren et al.~\cite{ren2022simple} proposed SSL based on a transformer using mixed images and synthesized representations.
The method improves performance by adding learning to maximize the similarity between the representations of mixed images and synthesized representations.
MixSiam~\cite{zhang2022mix} is an SSL method that uses mixed images and representations synthesized by a maximizing operation to learn more discriminative representations.
Although some SSL methods using synthesized representation have been proposed, the logic operability of representations by synthesis has not been studied in detail.

\subsection{Representation Operability}
Representation operations and control have been studied regarding image generation, image retrieval, interpretability of representations, and generation of new representations, among others~\cite{higgins2016beta, chen2016infogan, sun2021cross, shen2022mix, ren2022simple, lee2021imix, kim2020mixco, zhang2022mix, mirza2014conditional, pmlr-v139-ramesh21a, levy2024chatting, guo2018dialog}.
For example, $\beta$-VAE~\cite{higgins2016beta} and Guided-VAE~\cite{ding2020guided} improve the interpretability of representations by disentangling them, which has been found to be beneficial for both image retrieval and generation.
info-GAN~\cite{chen2016infogan} and DALL-E~\cite{pmlr-v139-ramesh21a} improve the performance of control of generated images by generating representations from controllable latent variables and languages.
In addition, there are also methods\cite{levy2024chatting, guo2018dialog} that involve modifying representations through interactions with humans to retrieve intended images.
Furthermore, i-mix~\cite{lee2021imix} and MixSiam~\cite{zhang2022mix} realize SSL using mixed images by generating new representations upon synthesizing the representations of images.
These methods enable representation learning for various features.
However, they fell short of enabling logic operations between representations. 

\section{PRELIMINARY}\label{sec:preliminary}
\subsection{Self-Supervised Learning}
\begin{figure}[t]
  \centering
  \includegraphics[width=0.85\linewidth]{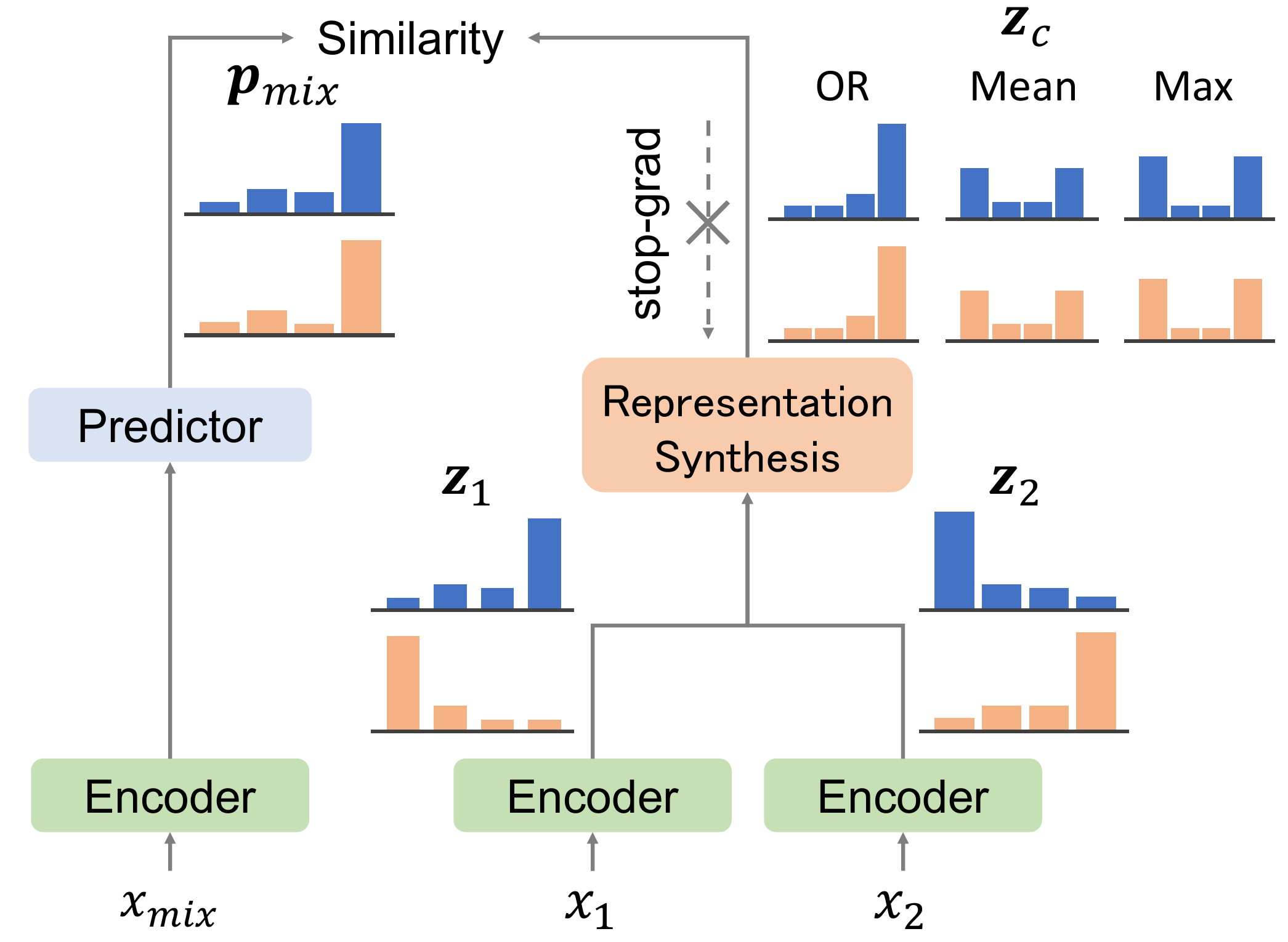}
  \caption{
  Overview of SSL using mixed images and representation synthesis, such as the logic (OR), mean, and maximum operations.
  }
  \label{fig:method}
\end{figure}
Our method is based on non-contrastive SSL, such as SimSiam~\cite{chen2021exploring} and BYOL~\cite{grill2020bootstrap}, DINO~\cite{caron2021emerging}.
The method learns representation prediction by minimizing the distance between the representations of images $x_i$ and ${x_i}'$ obtained from the random augmentation of image $X_i$.
Here, $i$ indicates the image index.
This method need two encoders $f_{\theta}$ and $g_{\phi}$ to predict image representations $\bm{z}_i,\ \bm{z}_i'$  and $\bm{p}_i,\ \bm{p}_i'$.
$\theta$ and $\phi$ are the parameters.

The loss $\mathcal{L}_{normal}$ is also defined by $\mathcal{D}(\cdot)$, which describe the difference between the representations, as follows;
\begin{align}
  \mathcal{L}_{normal} &= \frac{1}{2}\mathcal{D}(\bm{p}_i, \operatorname{stopgrad}(\bm{z}_i')) + \frac{1}{2}\mathcal{D}(\bm{p}_i', \operatorname{stopgrad}(\bm{z}_i)).  \label{eq:normal_loss}
\end{align}
The stop-gradient operation $\operatorname{stopgrad}(\cdot)$~\cite{chen2021exploring} prevents gradient calculation for the respective variable during backpropagation to avoid learning collapse.
The model is trained by $\underset{\phi}{\mathrm{min}}\ \mathcal{L}_{normal}$.
$\mathcal{D}(\cdot)$ differs depending on the style of representation, for example, negative cosine similarity~\cite{chen2021exploring, grill2020bootstrap} and cross entropy~\cite{caron2021emerging}.
\subsection{Self-Supervised Learning with Representation Synthesis}\label{sec:SSL_withsync}
This section describes a SSL method using the mixed image $x_{mix}$ and the synthesized representation $\bm{z}_{c}$.
Fig.~\ref{fig:method} shows an overview of the SSL method using mixed images and representation synthesis.
Mixup~\cite{zhang2018mixup} is used to generate the mixed image $x_{mix}$,  $\bm{z}_{c}$ is generated using representations $\bm{z}_1$ and $\bm{z}_2$ and the synthesis function $\rm{Mix}(\cdot)$.
$\lambda_{mix} \in [0,1]$ is the Mixup parameter.
\begin{align}
x_{mix} &= \lambda_{mix}x_1 + (1-\lambda_{mix})x_2, \\
\bm{z}_{c} &= \mathrm{Mix}(\bm{z}_1, \bm{z}_2)
\end{align}
We define another loss function $\mathcal{L}_{mix}$ to minimize the distance between $\bm{p}_{mix}$, which is the representation of $x_{mix}$ and the synthesized representation $\bm{z}_{c}$.
In addition, we also define the weighted average of $\mathcal{L}_{normal}$ and $\mathcal{L}_{mix}$ as the loss function $\mathcal{L}$.
The model is trained by $\underset{\phi}{\mathrm{min}}\ \mathcal{L}$.
$\alpha \in [0, 1]$ is a hyperparameter.
\begin{align}
  \mathcal{L}_{mix} &= \mathcal{D}(\bm{p}_{mix}, \operatorname{stopgrad}(\bm{z}_{c})), \label{eqn:loss_mix}\\
  \mathcal{L} &= \alpha\mathcal{L}_{normal} + (1-\alpha)\mathcal{L}_{mix} \label{eqn:loss}
\end{align}
\section{METHOD}\label{sec:method}
We propose a SSL method for logic-operable representations.
First, we describe a probabilistic extension of the many-valued logic and operations in Sec.~\ref{sec:method}-~\ref{logi_operation}.
Then, we describe SSL with representations which comprise multiple categorical distributions in Sec.~\ref{sec:method}-~\ref{sec:ssl_with_mcat}.
Afterward, we describe how to synthesize the representation using several operations in Sec.~\ref{sec:method}-~\ref{sec:proposed_method}.
\subsection{Logic Operation of the Probabilistic Many-Valued Logic}\label{logi_operation}
%
\begin{figure}[t]
    \centering
    \includegraphics[width=0.9\linewidth]{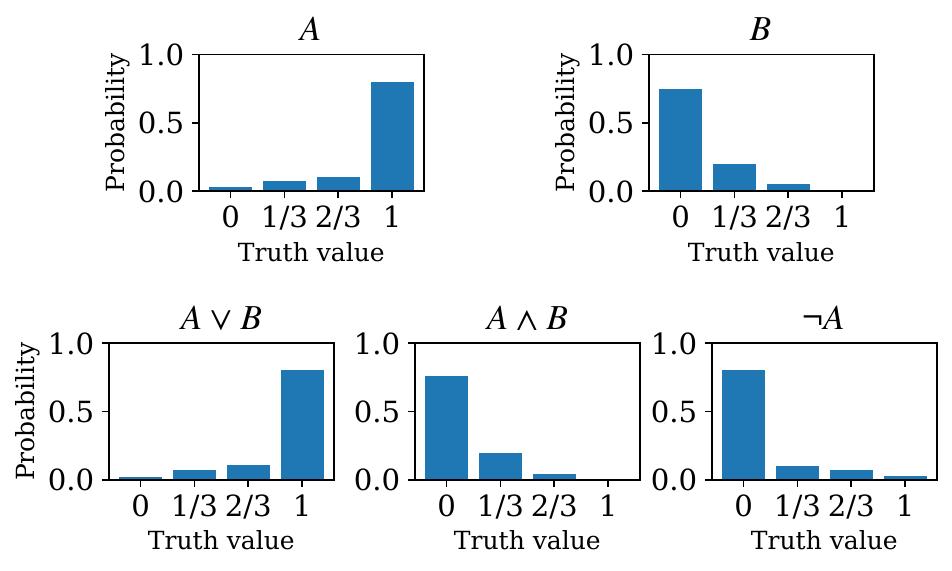}
    \caption{
        Example of the probability of each truth value for $n=4$.
        }
    \label{fig:synthesis_example}
\end{figure}
A many-valued logic refers to a logic system with truth values other than 0 and 1.
In this study, we considered G\"{o}del logics~\cite{godel1932intuitionistischen}.
In G\"{o}del logics, a family $G_{n}$ of many-valued logics possess $n$ truth values $0, \frac{1}{n-1}, \frac{2}{n-1},..., 1$.
Such truth values other than 0 and 1 realize the expression of uncertainty.
The logic operations of variables $A$ and $B$ are defined as follows:
$A$ and $B$ are the truth values, and $A, B \in [0,1]$.
\begin{align}
A &\lor B \coloneqq \max(A, B)
\label{eqn:or}\\
A &\land B \coloneqq \min(A, B)
\end{align}
We further define the NOT operation as $\neg A = 1 - A $.

When $A$ and $B$ follow categorical distributions, they are written as follows:
\begin{align}
    A &\sim \mathrm{Cat}(A | \mathbf{\pi_a}=\{a_{0}, a_{1}, ..., a_{n-1}\})
    \label{eqn:a_cat}\\
    B &\sim \mathrm{Cat}(B | \mathbf{\pi_b}=\{b_{0}, b_{1}, ..., b_{n-1}\})
    \label{eqn:b_cat}
\end{align}
where $a_i$ and $b_i$ represent probabilities.
We define $P(A=\frac{i}{n-1})=a_i$ and $P(B=\frac{i}{n-1})=b_i$.

Then, we discuss $P(A \lor B=\frac{i}{n-1})$, $P(A \land B=\frac{i}{n-1})$ and $P(\neg A=\frac{i}{n-1})$. 
They are described as;
\begin{align}
    P(A \lor B=\frac{i}{n-1}) &= a_{i}\sum_{j=0}^{i}b_{j} + b_{ i}\sum_{j=0}^{i}a_{j} - a_{i}b_{i},
    \label{eqn:p_or}\\
    P(A \land B=\frac{i}{n-1}) &= a_{i}\sum_{j=i}^{n-1}b_{j} + b_{ i}\sum_{j=i}^{n-1}a_{j} - a_{i}b_{i},
    \label{eqn:p_land} \\
    P(\neg A=\frac{i}{n-1}) &= a_{n-1-i}
    \label{eqn:p_neg}
\end{align}
Fig.~\ref{fig:synthesis_example} shows an example of the probability of each truth value of $A$, $B$, $A \lor B$, $A \land B$ 
 and $\neg A$ for $n=4$.
%
%
\subsection{Self-Supervised Learning with Multiple Categorical Distributions}\label{sec:ssl_with_mcat}
In this section, we describe SSL with representations which comprise multiple categorical distributions.
We prepare encoder $f_{\theta}$ and predictor $h_{\psi}$.
The second encoder $g_{\phi}$ is defined as $g_{\phi} = h_{\psi} \circ f_{\theta}$.
$\theta$, $\psi$ and $\phi$ are the parameters.
These encoders predict representations as follows;
\begin{align}
    \bm{\tilde{z}}_i = f_{\theta}(x_i)&,\  \bm{\tilde{z}}_i' = f_{\theta}(x_i') \label{eqn:z_1},\\
    \bm{\tilde{p}}_i = g_{\phi}(x_i)&,\  \bm{\tilde{p}}_i' = g_{\phi}(x_i') \label{eqn:p_1}.
\end{align}
Representations $\bm{\tilde{z}}_i$ and $\bm{\tilde{p}}_i$ comprise $N$ $M$-dimensional vectors with a total number of dimensions of $N \times M$.
Each vector is defined as $\bm{\tilde{z}}_{i, j}$, $\bm{\tilde{p}}_{i, j}$ $(j = 1, 2,..., N)$.
Consequently, the vectors $\bm{\tilde{z}}_{i, j}$, $\bm{\tilde{p}}_{i, j}$ are transformed into categorical distributions using the softmax function.
\begin{align}
    \bm{z}_{i, j} &= \mathrm{softmax}\bigl((\bm{\tilde{z}}_{i, j}-c)/\tau_{t}\bigr) \label{eqn:z_2}\\
    \bm{p}_{i, j} &= \mathrm{softmax}\bigl(\bm{\tilde{p}}_{i, j}/\tau_{s}\bigr) \label{eqn:p_2}
\end{align}
$\bm{z}_{i, j}'$ and $\bm{p}_{i, j}'$ are also obtained by replacing $\bm{\tilde{z}}_{i, j}$ and $\bm{\tilde{p}}_{i, j}$ in Eqs.~(\ref{eqn:z_2}) and (\ref{eqn:p_2}) with $\bm{\tilde{z}}_{i, j}'$ and $\bm{\tilde{p}}_{i, j}'$ respectively.
Representations $\bm{z}_i$ and $\bm{p}_i$ comprise $N$ $M$-dimensional categorical distributions.
Thus, $\bm{z}_{i, j},\ \bm{p}_{i, j} \in \mathbb{R}^M$ and $\bm{z}_i,\ \bm{p}_i \in \mathbb{R}^{N \times M}$.
The $k$-th class parameter of the $j$-th categorical distribution is described by $z_{i, j, k},\ p_{i, j, k} \in [0,1]$.
$\tau_{t}$ and $\tau_{s}$ are the parameters of sharpening, and $c$ is a parameter of centering\cite{caron2021emerging}.

By considering cross-entropy, $\mathcal{D}(\cdot)$ in Eq.~(\ref{eq:normal_loss}) is defined as follows\footnote{When $N=1$, Eq.~(\ref{eqn:cross-entropy}) is equivalent to that of DINO~\cite{caron2021emerging}.};
\begin{align}
    \mathcal{D}(\bm{p}_i, \operatorname{stopgrad}(\bm{z}_i')) &= -\frac{1}{N}\sum_{j=1}^{N}\sum_{k=1}^{M}{z_{i, j, k}'\log{p_{i, j, k}}}. \label{eqn:cross-entropy}
\end{align}
\subsection{Representation Synthesis}\label{sec:proposed_method}
In this section, we explain the operations used to synthesize representations, $\rm{Mix}(\cdot)$.
First, we describe the synthesis of representations using the mean and maximum operations used in existing methods.
Then, we describe representation synthesis using a many-valued logic operation to learn representations that are operable by logic operations.
%
\paragraph*{Representation Synthesis by the Mean Operation}
This synthetic method is referenced from SDMP~\cite{ren2022simple}.
The synthesized representation $\bm{z}_{c}$ is derived from $\bm{z}_1$, $\bm{z}_2$, and the Mixup parameter $\lambda_{mix}$.
\begin{align}
\bm{z}_{c} = \lambda_{mix} \bm{z}_1 + (1 - \lambda_{mix}) \bm{z}_2
\end{align}
\paragraph*{Representation Synthesis by the Maximum Operation}
This synthetic method was referenced from MixSiam~\cite{guo2021mixsiam}.
The synthesized representation $\bm{z}_{c}$ is the normalized element-wise maximum of the prenormalized representations $f(x_1)$ and $f(x_2)$.
$\bm{z}_{c}$ is derived as follows, where the $\mathrm{maximum}$ means the elementwise maximum function;
%
%
\begin{align}
\bm{z}_{c, j} = \mathrm{softmax}\biggl(\frac{\mathrm{maximum}\bigl(\bm{\tilde{z}}_{1,j}, \bm{\tilde{z}}_{2,j}\bigr)-c}{\tau_{t}}\biggr).
\end{align}
\paragraph*{Representation Synthesis by Logic Operation}
We propose a representation synthesis method based on many-valued logics.
In our method, the degree of possession of each feature in the image is indicated by truth values to achieve logic operations between representations.
The representation $\bm{z}_c$ in Eq.~(\ref{eqn:loss_mix}) is synthesized by a logic operation.

\textbf{Feature-possession degree}:\ \ 
First, we describe how to express each feature-possession degree of an image.
$Z_{i,j}$ and $P_{i,j}$ are stochastic variables that follow the $j$-th categorical distribution of $\bm{z}_{i}$, $\bm{p}_{i}$.
When $Z_{i,j}$ and $P_{i,j}$ follow G\"{o}del logics and have $M$ truth values ($0, \frac{1}{M-1}, ..., 1$), such as Eq.~(\ref{eqn:a_cat}) and (\ref{eqn:b_cat}), $Z_{i,j}$ and $P_{i,j}$ are described as follows:
\begin{align}
Z_{i, j} \sim \mathrm{Cat}(Z_{i, j} | \mathbf{\pi}=\{ z_{i, j, 1},  z_{i, j, 2},..., z_{i, j, M}\}),
\label{eqn:z_cat}\\
P_{i, j} \sim \mathrm{Cat}(P_{i, j} | \mathbf{\pi}=\{ p_{i, j, 1},  p_{i, j, 2},..., p_{i, j, M}\}),
\label{eqn:p_cat}\\
P(Z_{i, j}=\frac{k-1}{M-1}) =  z_{i, j, k}.
\label{eqn:prob_z}\\
P(P_{i, j}=\frac{k-1}{M-1}) =  p_{i, j, k}.
\label{eqn:prob_p}
\end{align}
We assume that each categorical distribution $Z_{i, j}$ corresponds to a feature of image.
In other words, the larger the expected value of $Z_{i, j}$, the more features of the image corresponding to $Z_{i, j}$ that it possesses.
$P_{i, j}$ also has same properties.

\textbf{Representation Synthesis by Logic Operation}:\ \
Logic operation is possible because each feature-possession degree is expressed as a truth value.
Here, we explain the OR, AND, and NOT operations.
$Z_{1,j}$ and $Z_{2,j}$ are the truth values that follow the $j$-th categorical distribution of $\bm{z}_{1}$, $\bm{z}_{2}$.
We set the truth values after the operation of these truth values as follows:
\begin{align}
    Z_{\mathrm{OR}, j} \coloneqq Z_{1, j} \lor Z_{2, j} \\
    Z_{\mathrm{AND}, j} \coloneqq Z_{1, j} \land Z_{2, j} \\
    Z_{\mathrm{NOT}, j} \coloneqq \neg Z_{1, j}
\end{align}
%
$Z_{1,j}$ and $Z_{2,j}$ follow categorical distributions as shown in Eq.~(\ref{eqn:prob_z}).
From Eqs.~(\ref{eqn:p_or}), (\ref{eqn:p_land}) and (\ref{eqn:p_neg}), $z_{\mathrm{OR}, j, k}$, $z_{\mathrm{AND}, j, k}$, and $z_{\mathrm{NOT}, j, k}$ are as follows:
\begin{align}
    &z_{\mathrm{OR}, j, K} = P(Z_{1, j} \lor Z_{2, j}=\frac{K-1}{M-1}) \nonumber \\
    &= z_{1, j, K}\sum_{k=1}^{K}z_{2, j, k} + z_{2, j, K}\sum_{k=1}^{K}z_{1, j, k} - z_{1, j, K}z_{2, j, K} \label{eqn:or_operation} \\
    &z_{\mathrm{AND}, j, K} = P(Z_{1, j} \land Z_{2, j}=\frac{K-1}{M-1}) \nonumber \\
    &= z_{1, j, K}\sum_{k=K}^{M}z_{2, j, k} + z_{2, j, K}\sum_{k=K}^{M}z_{1, j, k} - z_{1, j, K}z_{2, j, K} \label{eqn:and_operation}
\end{align}
\begin{align}
    z_{\mathrm{NOT}, j, K} &= P(\neg Z_{1, j}=\frac{K-1}{M-1}) \nonumber \\
    &= z_{1, j, M-K+1} \label{eqn:not_operation}
\end{align}
%
We define $\bm{z}_1 \lor \bm{z}_2$ as operations that perform Eq.~(\ref{eqn:or_operation}) and $\bm{z}_1 \land \bm{z}_2$ as operations that perform Eq.~(\ref{eqn:and_operation}) for all $j$, $k$ in $\bm{z}_1,\ \bm{z}_2$.
In addition, we define $\neg \bm{z}_1$ as operations that perform Eq.~(\ref{eqn:not_operation}) for all $j$, $k$ in $\bm{z}_1$.
These operations can also be performed on $\bm{p}_1$ and $\bm{p}_2$.

\textbf{Synthesization of $\bm{z}_c$}:\ \ 
As the representation $\bm{p}_{mix}$ is considered to possess the features of images $x_1$ and $x_2$, the synthesized representation $\bm{z}_c$ is designed such that it has features of $\bm{z}_1$ and $\bm{z}_2$.
To satisfy this requirement, we set $\bm{z}_{c} = \bm{z}_{1} \lor \bm{z}_{2}$.
%
\subsection{Expected Value Loss}\label{sec:exp_loss}
This section describes a new loss considering the expected value of the feature-possession degree.
For example, when there are three stochastic variables of the truth value,
\footnotesize
\begin{align}
    Z_{1, j} &\sim \mathrm{Cat}(Z_{1, j} | \mathbf{\pi}=\{z_{1, j, 1}=0.8,  z_{1, j, 2}=0.1, z_{1, j, 3}=0.1\}), \nonumber \\
    P_{2, j} &\sim \mathrm{Cat}(P_{2, j} | \mathbf{\pi}=\{p_{2, j, 1}=0.1,  p_{2, j, 2}=0.8, p_{2, j, 3}=0.1\}), \nonumber \\
    P_{3, j} &\sim \mathrm{Cat}(P_{3, j} | \mathbf{\pi}=\{p_{3, j, 1}=0.1,  p_{3, j, 2}=0.1, p_{3, j, 3}=0.8\}), \nonumber
\end{align}
\normalsize
the cross-entropy loss of $Z_{1, j}$ and $P_{2, j}$ is the same as that of $Z_{1, j}$ and $P_{3, j}$, though the expected truth values of $P_{2, j}$ and $P_{3, j}$ are different.
To facilitate learning the appropriate feature-possession degrees, we introduce the new loss which penalizes more when the difference between the expected values of feature-possession degrees is large.
By considering the binary cross-entropy of the expected values of the random variables $Z_{i, j}$ and $P_{i, j}$, it is described as follows;
\begin{align}
    \mathrm{BCE}(\bm{p}_{i, j}, \bm{z}_{i, j}) &= \nonumber \\ 
    - E[Z_{i, j}]& \log{E[P_{i, j}]} - (1-E[Z_{i, j}]) \log{(1-E[P_{i, j}])}.
\end{align}
$E[\cdot]$ indicates the expected value.
The loss functions $\mathcal{L}_{exp}$ is defined as follows;
\begin{align}
    \mathcal{L}_{1} &= \frac{1}{2N}\sum_{j=1}^{N} \bigl(\mathrm{BCE}\bigl(\bm{p}_{i, j}, \operatorname{stopgrad}(\bm{z}_{i, j}')) \nonumber \\ 
    &\ \ \ \ \ \ \ \ \ \ \ \ \ \ \ \ \ + \mathrm{BCE}(\bm{p}_{i, j}', \operatorname{stopgrad}(\bm{z}_{i, j})\bigr)\bigr), \\
    \mathcal{L}_{2} &= \frac{1}{N}\sum_{j=1}^{N} \mathrm{BCE}\bigl(\bm{p}_{mix, j}, \operatorname{stopgrad}(\bm{z}_{c, j})), \\
    \mathcal{L}_{exp} &= \alpha \mathcal{L}_{1} + (1-\alpha)\mathcal{L}_{2}. \label{eqn:exp_loss_base}
\end{align}
Finally, we use the weighted average of $\mathcal{L}$ from Eq.~(\ref{eqn:loss}) and $\mathcal{L}_{exp}$ as the loss function.
$\beta \in [0, 1]$ is a hyperparameter.
\begin{align}
    \label{eqn:exp_loss}
    \mathcal{L}_{prop} = \beta \mathcal{L}_{exp} &+ (1-\beta) \mathcal{L} \\
    \underset{\phi}{\mathrm{min}} & \ \mathcal{L}_{prop}
\end{align}

\section{EXPERIMENT}\label{sec:exp}
In this section, we experiment with the proposed method from several perspectives.
First, we compare the performance of the proposed and existing methods on classification tasks. (Sec.~\ref{sec:exp}-\ref{sec:single_classification}, \ref{sec:exp}-\ref{sec:multi_classification}).
Further, we confirm the relationship between the feature-possession degree and images and show that the feature-possession degree is learned as intended (Sec.~\ref{sec:exp}-\ref{sec:posession-degree}).
In addition, we confirm the operability of the representation via logic operations using image retrieval (Sec.~\ref{sec:exp}-\ref{sec:image_retrieval}).

\subsection{Implementation}
We implemented four methods: vanilla and representation synthesis by using the mean, maximum, and logic operations, as described in Sec.~\ref{sec:preliminary}-\ref{sec:proposed_method}.
Vanilla is a method that does not use mixed images or representation synthesis.
Unless stated otherwise, the setup for each method is as follows.

\textbf{Experimental settings for networks}\ \ 
The projector of the encoder $f$ and predictor $h$ of the proposed method are the same as those of SimSiam~.
The backbone network of the encoder $f$ is Resnet-18~\cite{he2016deep}.

\textbf{Experimental settings for pretraining}\ \ 
Unless stated otherwise, the number of categorical distributions within the representation $N$ is 512, and the number of classes in each categorical distribution $M$ is 8.
The total number of dimensions of the representation, $N \times M$, is 4096.
We used momentum-SGD for pretraining. 
The learning rate was 0.027, and SGD momentum was 0.9. 
The learning rate followed a cosine-decay schedule.
The weight decay was 1e-4.
The batch size was 32.
The parameters of sharpening were $\tau_{s}=0.1$ and $\tau_{t}=0.04$.
The parameter of centering $c$ was updated as in DINO, with the update parameter set to $m=0.9$.
The augmentation settings were the same as those used in SimSiam.
The image-mixing method was Mixup~\cite{zhang2018mixup}. 
The parameter for Mixup, $\lambda_{mix}$, was set to 0.5.
The loss weight $\alpha$ in Eq.~(\ref{eqn:loss}) and (\ref{eqn:exp_loss_base}) is 0.5.
$\beta$, the parameter for the expected value loss of the logic operation method in Eq. (\ref{eqn:exp_loss}) is 0.6.

\subsection{Single-Label Image Classification}\label{sec:single_classification}
\begin{table}[t]
 \caption{
    \textbf{Top-1 accuracy of linear-evaluation}
    For each method, eight experiments were performed, and the mean and standard deviation (SD) of the Top-1 accuracy were calculated.
    The results are expressed as the mean $\pm$ SD.
    }
 \label{table:result}
 \centering
 \scalebox{0.8}{
  \begin{tabular}{lrr}
   \hline
   Method & ImageNet100 & Cifar10 \\
   \hline \hline
   Vanilla & \third{72.90}{0.33}  & \third{89.83}{0.30}\\
   Mean operation~\cite{ren2022simple}  & \third{74.44}{1.64}  & \third{91.24}{0.32} \\
   Maximum operation~\cite{guo2021mixsiam}  & \best{75.41}{0.85}  & \third{90.72}{0.18}  \\
   \hline
   \rowcolor{black!10!} Logic operation ($N=2048,\ M=2$) & \third{72.86}{2.63} & \third{90.96}{0.14} \\
   \rowcolor{black!10!} Logic operation ($N=1024,\ M=4$) & \third{74.00}{2.74} & \third{90.90}{0.72} \\
   \rowcolor{black!25!} Logic operation ($N=512,\ M=8$) & \best{75.41}{0.95} &  \best{91.28}{0.22} \\
   \rowcolor{black!10!} Logic operation ($N=256,\ M=16$) & \third{73.59}{0.53} & \third{90.79}{0.24} \\
   \rowcolor{black!10!} Logic operation ($N=128,\ M=32$) & \third{73.74}{0.38} & \third{90.69}{0.49} \\
   \hline
  \end{tabular}
  }
\end{table}
We experimented with single-label image classification to verify the effectiveness of our method.
First, we performed self-supervised pretraining with ImageNet100~\cite{deng2009imagenet}\footnote{ImageNet100 is a 100-category subset of ImageNet~\cite{deng2009imagenet}.} and CIFAR10~\cite{krizhevsky2009learning} dataset without labels to learn image representations. 
Then, we trained a linear classifier on frozen representations on the training set of the dataset with the labels. 
Finally, we calculated the top-1 accuracy of the test set and used it as an evaluation metric.

Now, we describe the experimental setup for learning a linear classifier.
We used LARS~\cite{you2017large} as an optimizer for linear evaluation. 
The learning rate was 1.6, and the SGD momentum was 0.9. 
The weight decay was 0.0, and the batch size was 512. 
The image augmentation was performed as in SimSiam.
The number of epochs was 200.
We used the model with the highest top-1 accuracy as the validation set for testing.
When we trained it, we cropped an image randomly and resized it to $224 \times 224$ before inputting it into the model.
The area of a randomly cropped image ranged from 0.08 to 1.0 of the area of the original image.
In the validation and test phases, the image was resized to $256 \times 256$, and the center was cropped to $224 \times 224$.

Tab.~\ref{table:result} summarizes the results.
We compared four methods: Vanilla and representation synthesis using the mean, maximum, and logic operations.
Our method performs competitively compared with existing synthetic methods.
In the logic operations, we compared $M$, the number of truth values of the many-valued logic, from 2 to 16.
The results of ImageNet100 showed that the many-valued logic $(M > 2)$, which can express the uncertainty of the feature-possession degree, improved accuracy compared to the binary case $(M = 2)$.
The performances are lower for $M=16$ and $32$ than for $M=8$.
We consider this is because if $M$ is increased under the condition that $NM$ is constant, $N$ , the number of the feature-possession degree, decreases and a sufficient variety of feature-possession degrees cannot be described by a representation.
\subsection{Multilabel Image Classification}\label{sec:multi_classification}
\begin{table}[t]
 \caption{
    \textbf{Results of multilabel classification}
    For each method, eight experiments were performed, and the mean and SD of mAP were calculated.
    The results are expressed as the mean $\pm$ SD.
    }
 \label{table:result_multilabel}
 \centering
 \scalebox{0.8}{
  \begin{tabular}{lrr}
   \hline
   Method & PascalVOC & COCO \\
   \hline \hline
   Vanilla & \third{61.66}{0.31}& \third{43.76}{0.29}\\
   Mean operation~\cite{ren2022simple}  & \third{61.93}{0.56}& \third{44.05}{0.33}\\
   Maximum operation~\cite{guo2021mixsiam}  & \third{61.19}{0.32}& \third{43.86}{0.23}\\
   \hline
   \rowcolor{black!10!} Logic operation ($N=2048,\ M=2$) & \third{60.60}{0.99}& \third{43.59}{0.89}\\
   \rowcolor{black!10!} Logic operation ($N=1024,\ M=4$) & \third{61.24}{1.53}& \third{43.80}{1.06}\\
   \rowcolor{black!25!} Logic operation ($N=512, \ M=8$) & \best{62.26}{0.32}& \best{44.59}{0.31}\\
   \rowcolor{black!10!} Logic operation ($N=256,\ M=16$) & \third{61.47}{0.50}& \third{44.11}{0.33}\\
   \rowcolor{black!10!} Logic operation ($N=128, M=32$) & \third{61.85}{0.29}& \third{43.86}{0.26}\\
   \hline
  \end{tabular}
  }
\end{table}
We also experimented with multilabel image classification to verify the effectiveness of our method.
The model pre-trained at 200 epochs by ImageNet100 was trained on the multilabel classification dataset and evaluated in terms of mean average precision (mAP).

In the experiments, the backbone network was fixed, and only the final fully-connected layer was trained.
The number of epochs was 500, and the learning rate was 0.02.
When we trained the models, the input image was resized to 640 × 640, and we randomly selected \{640, 576, 512, 384, 320\} as the width and height to crop the images.
Finally, the cropped images were resized to 224 × 224 pixels. 
When we evaluated the models, we resized the input image to 256 × 256 and performed center-crop operation with a size of 224 × 224.

We used Microsoft COCO~\cite{lin2014microsoft} and PascalVOC~\cite{everingham2010pascal} as multilabel classification datasets, which contain 80 and 20 categories, respectively.
When using COCO, we trained models using the training set and evaluated them using the validation set because ground truth annotations for the test set were unavailable.
When using PascalVOC, we trained models using the trainval2007 and trainval2012 sets and evaluated them using the test2007 set.

Table.~\ref{table:result_multilabel} shows the results.
Our methods are competitive with other representation synthesis methods.
\subsection{Evaluation of the Feature-Possession Degree}\label{sec:posession-degree}
\begin{figure}[t]
    \centering
    \includegraphics[width=0.9\linewidth]{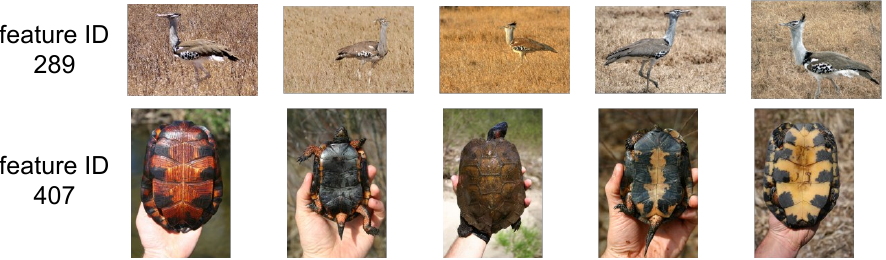}
    \caption{
        \textbf{The-5 images with the highest feature-possession degrees.}
        Feature ID is the index number of feature-possession degrees.
        }
    \label{fig:example_top_images}
\end{figure}
%
%
In this section, we verify that the feature-possession degree $E[P_{i, j}]$ corresponds to a certain image feature and is learned as hypothesized.
Fig.~\ref{fig:example_top_images} shows the examples of the top five images with some feature-possession degrees.
We used ImageNet100 dataset.
This results suggest that the feature-possession degree is learned as an indicator of how much of a certain feature an input image has because images with large feature-possession degrees tend to be similar.

%
\begin{figure}[t]
  \centering
  \includegraphics[width=0.55\linewidth]{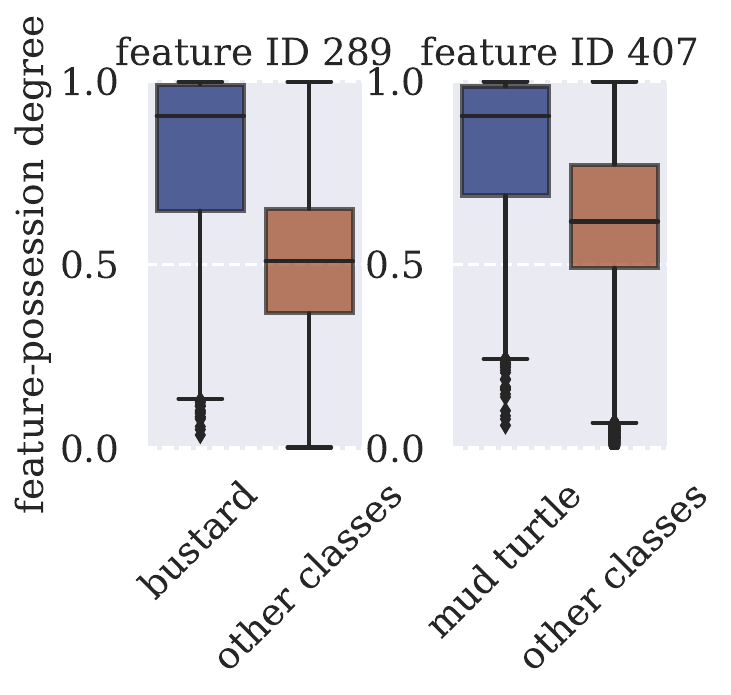}
  \caption{
  Boxplot of the feature-possession degrees of particular classes compared with others.
  }
  \label{fig:boxplot_degree}
\end{figure}
Next, we investigated the relationship between the feature-possession degree and the image class.
Fig.~\ref{fig:boxplot_degree} shows the boxplot of a particular feature-possession degree in one class and the other.
For example, the representation of class “bustard” images has a 289th feature-possession degree larger than the representations of the other classes.
The result of the larger feature-possession degree for the images of a particular class indicates that the feature-possession degree corresponds to the image feature that is frequently included in that class.

\begin{figure}[t]
    \centering
    \includegraphics[width=0.95\linewidth]{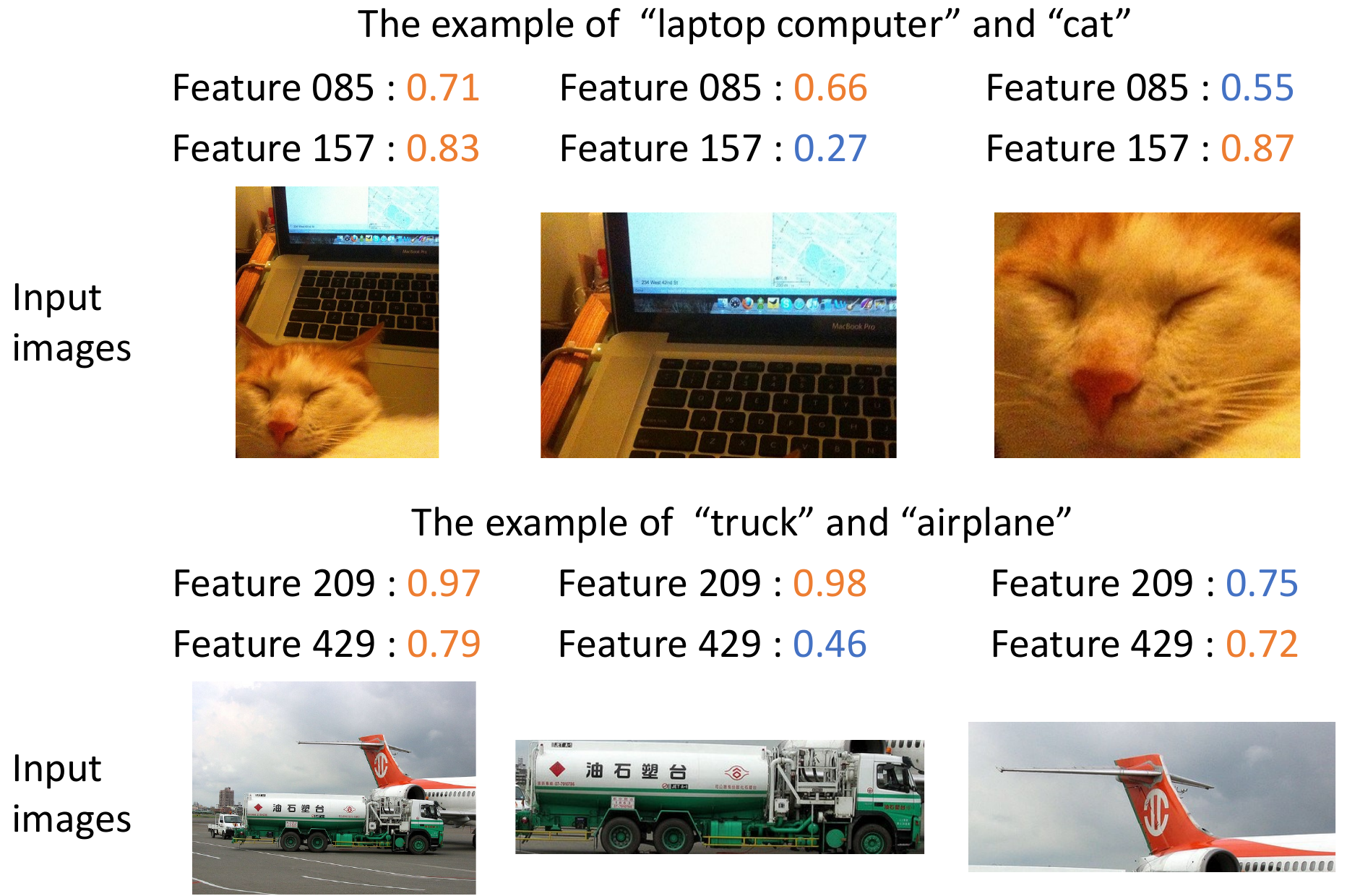}
    \caption{
        \textbf{Input images and two feature-possession degrees.}
        The left image contains two objects, both with feature-possession degrees, which are relatively high.
        Meanwhile, the middle and right cropped images, which have only one object, show that one feature-possession degree is relatively small.
        }
    \label{fig:degee_examples_multilabel}
\end{figure}

Using a multilabel classification dataset, we also investigated the relationship between the feature-possession degree and the objects.
Fig.~\ref{fig:degee_examples_multilabel} shows examples of the relationship.
The left image contains two objects; the middle and right images contain one object cropped from the left image.
We also show two feature-possession degrees in these images.
Both feature-possession degrees are relatively high in the left image containing both objects.
However, in the middle and right images, one is relatively high, and the other is relatively low, indicating there is a relationship between the objects in the image and the feature-possession degree.
%
%
\subsection{Image Retrieval from Synthesized Representations}\label{sec:image_retrieval}
%
Here, we discuss the operability between representations by performing image retrieval using MNIST and PascalVOC datasets.

\subsubsection{MNIST}\label{sec:exp_mnist}
\begin{figure}[t]
    \centering
    \includegraphics[width=0.9\linewidth]{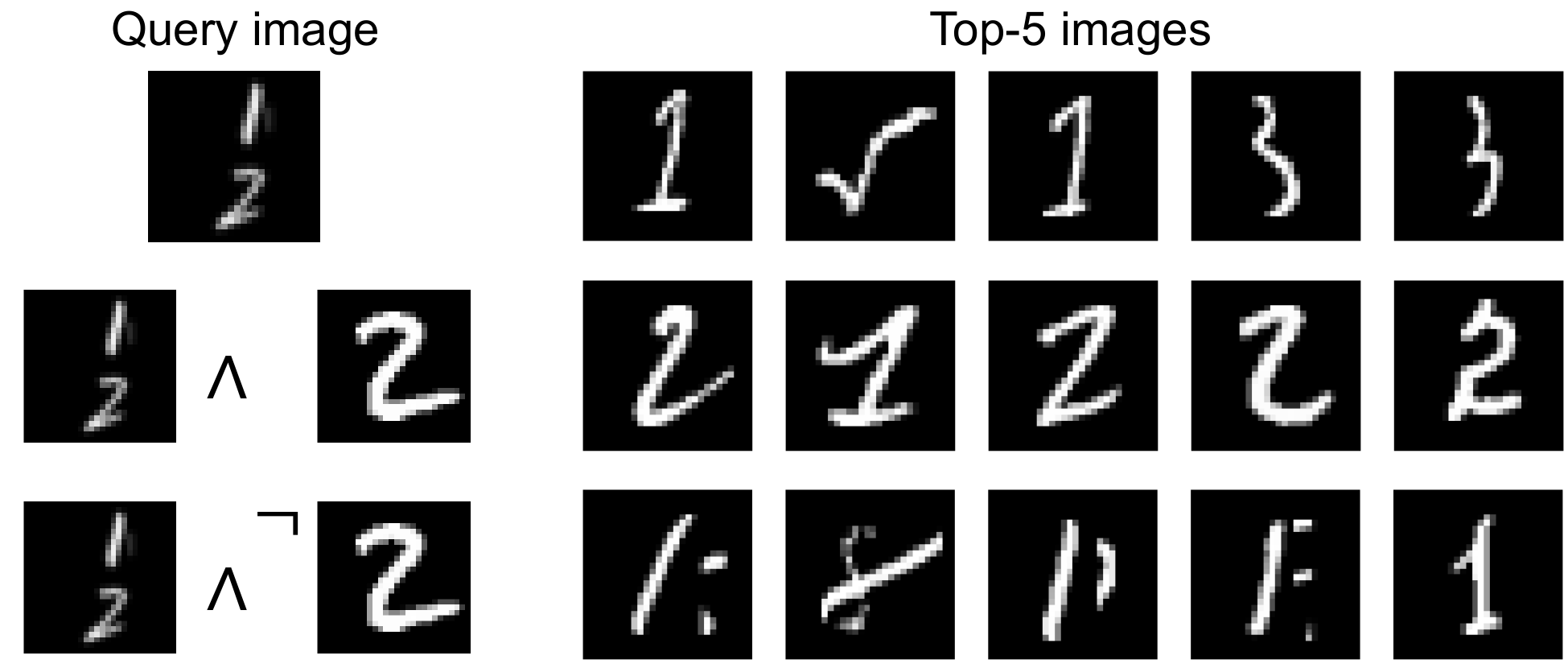}
    \caption{
        \textbf{Examples of image retrieval result by the representation operated by logic operation.}
        The top row shows images retrieved using a query for images containing both "1" and "2". 
        The middle row shows the results of an AND operation between the representations of images containing both '1' and '2' and those containing only '2', revealing an increase in images featuring only '2' compared to the top row. 
        }
    \label{fig:image_retrieval_mnist_example}
\end{figure}

First, we discuss the results of image retrieval using the MNIST dataset.
Fig.~\ref{fig:image_retrieval_mnist_example} shows the top five images of the image retrieval at each query representation.
The second row shows the result when the AND operation is performed on the representation of the image containing only ``2'' for the representation of the image containing both ``1'' and ``2''.
Since most of the top five images are class ``2'' images, we assume that the ``2'' feature remained in the representation of the query owing to the AND operation.
Conversely, when the negation of the representation of the class ``2'' image is performed AND operation, the features of ``2'' are removed from the query, indicating that the images of ``1'' appear more frequently in the top-5.
We used the MNIST training set as the gallery set, and query images were selected from the MNIST test set.
We used Eq.~(\ref{eqn:cross-entropy}) as the distance between two representations. 

In addition, we also quantitatively evaluated representation operability using an image retrieval task.
Let the set of classes for MNIST be defined as $L$, and given classes $a, b \in L$, the sets of images in the test set that contain only class $a$ or class $b$ are defined as $X_{\{a\}}$ and $X_{\{b\}}$.
Also, let $X_{\{a, b\}}$ denote the set of images mixup of images $x^{\{a\}}_i \in X_{\{a\}}$ and $x^{\{b\}}_j \in X_{\{b\}}$.
We define $\bm{p}^{\{a\}}_i$ as the representation obtained by inputting $x^{\{a\}}_i \in X_{\{a\}}$ into the pretrained model with reference to Eqs.~(\ref{eqn:z_1}), (\ref{eqn:p_1}), and (\ref{eqn:p_2}).
We used the top-50 occurrence rates of the class $a$ images as the evaluation value.
First, we prepared the evaluation values when $\bm{p}^{\{a, b\}}_i$, $\bm{p}^{\{a\}}_i$, and $\bm{p}^{\{b\}}_i$ are queries, and then used these evaluation values as the reference values.
Then, we prepared a representation expected to be close to the above three representations by the operations among the representations.
We derived the evaluation values when the representations generated by the operations were used as queries.
Operability is evaluated by comparing them with the reference values.
We used the model pretrained by MNIST by 500 epochs to predict representations.

The backbone network is composed of 3 convolution layers and 2 fully-connected layers.
The number of dimensions of the representation was 1024, the number of categorical distributions within the representations $N$ was 128, and the number of classes in each categorical distribution $M$ was 8.
We randomly selected 30 $x^{\{a\}}_i$, $x^{\{b\}}_i$, and $x^{\{a, b\}}_i$ from $X^{\{a\}}_i$, $X^{\{b\}}_i$, and $X^{\{a, b\}}_i$ and show the average evaluation value when those representations are the queries.
Moreover, we show the average of evaluation values of 900 combinations of two representations in operations such as $\bm{p}^{\{a\}}_i \lor \bm{p}^{\{b\}}_i$.

\begin{figure}[t]
  \begin{minipage}{\linewidth}
    \centering
    \includegraphics[width=0.99\linewidth]{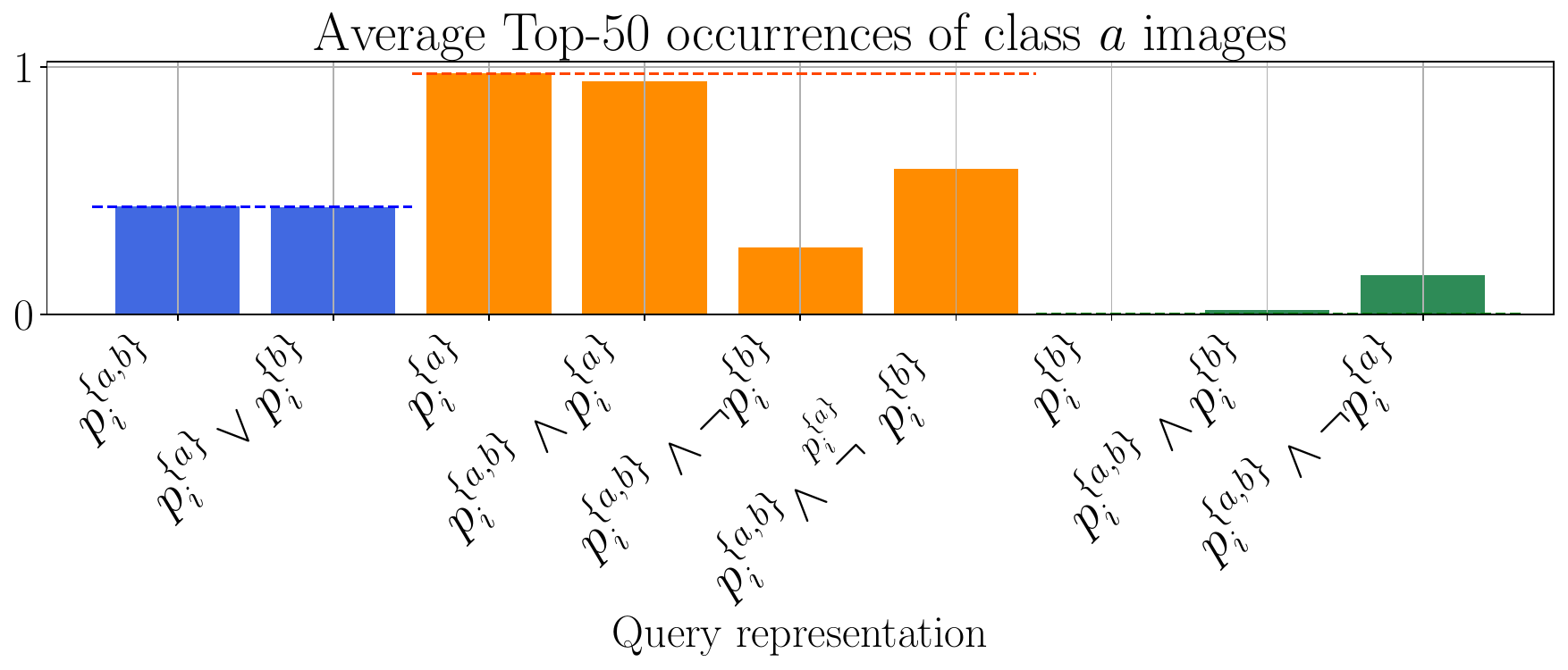}
    \subcaption{MNIST dataset}
    \label{fig:image_retrieval_rerult_rate}
  \end{minipage}
  \begin{minipage}{\linewidth}
    \centering
    \includegraphics[width=0.99\linewidth]{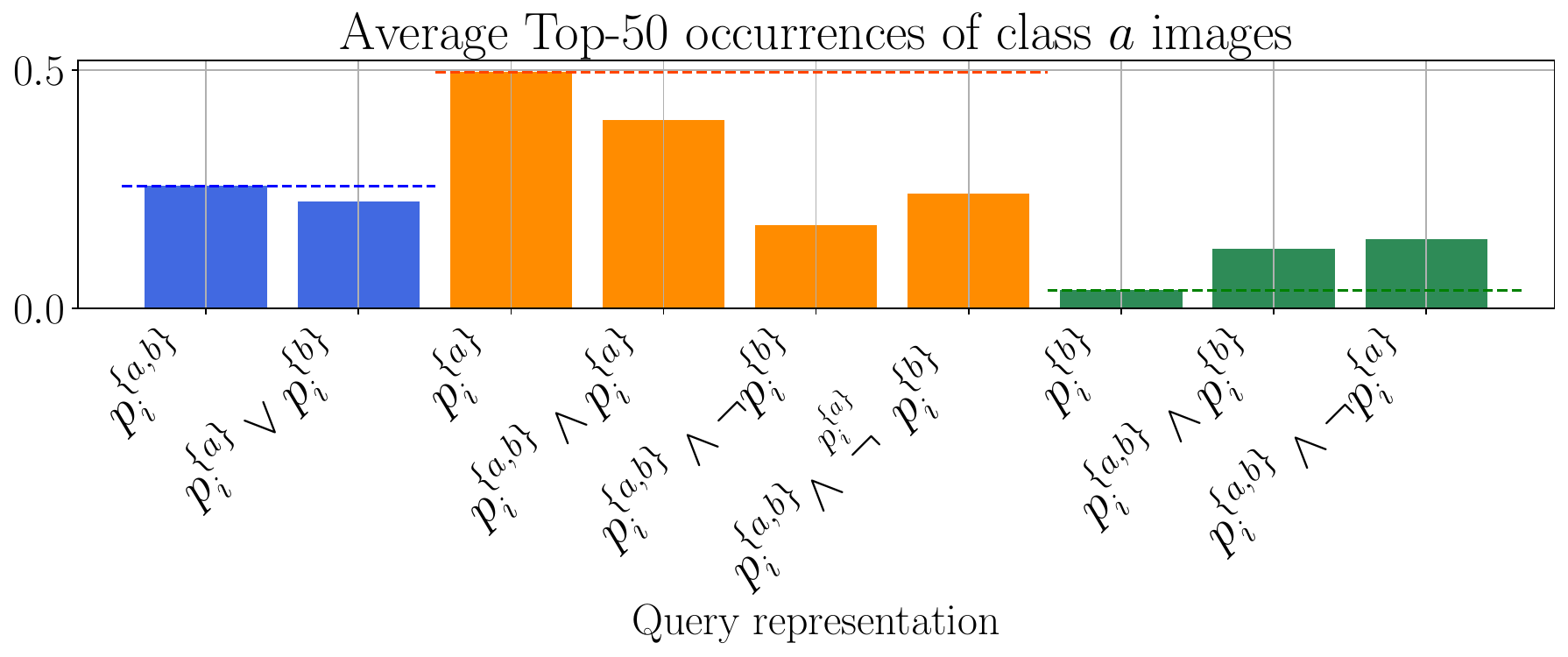}
    \subcaption{PascalVOC dataset}
    \label{fig:image_retrieval_rerult_rate_voc}
  \end{minipage}
  \caption{
    The occurrence rate of images of class $a$ in the top-50 when image retrieval is performed using $\bm{p}^{\{a, b\}}_i$, $\bm{p}^{\{a\}}_i$ and $\bm{p}^{\{b\}}_i$ and the representation after logic operation of them as queries.
    }
\end{figure}

Fig~\ref{fig:image_retrieval_rerult_rate} shows the average of the top-50 occurrence rates.
First, we compare $\bm{p}^{\{a, b\}}_i$ and $\bm{p}^{\{a\}}_i \lor \bm{p}^{\{b\}}_i$ (shown in blue).
Both are expected to have features of both classes $a$ and $b$.
The results show that the occurrence rates are very close, indicating that the OR operation works as intended.
Next, we compared $\bm{p}^{\{a, b\}}_i \land \bm{p}^{\{a\}}_i$ and $\bm{p}^{\{a, b\}}_i \land \neg \bm{p}^{\{b\}}_i$ with $\bm{p}^{\{a\}}_i$ (shown in orange), and $\bm{p}^{\{a, b\}}_i \land \bm{p}^{\{b\}}_i$ and $\bm{p}^{\{a, b\}}_i \land \neg \bm{p}^{\{a\}}_i$ with $\bm{p}^{\{b\}}_i$ (shown in green).
$\bm{p}^{\{a, b\}}_i \land \bm{p}^{\{a\}}_i$ and $\bm{p}^{\{a, b\}}_i \land \neg \bm{p}^{\{b\}}_i$ would be expected to have only class $a$ features, while $\bm{p}^{\{a, b\}}_i \land \bm{p}^{\{b\}}_i$ and $\bm{p}^{\{a, b\}}_i \land \neg \bm{p}^{\{a\}}_i$ would be expected to have class $b$ features.
The results show that $\bm{p}^{\{a, b\}}_i \land \bm{p}^{\{a\}}_i$ and $\bm{p}^{\{a, b\}}_i \land \bm{p}^{\{b\}}_i$ are very close to the reference values, indicating that the AND operation works as intended.
On the other hand, $\bm{p}^{\{a, b\}}_i \land \neg \bm{p}^{\{b\}}_i$ and $\bm{p}^{\{a, b\}}_i \land \neg \bm{p}^{\{a\}}_i$ is not close to the reference values.
We note that the relationship between a class and a feature-possession degree is not a direct one-to-one correspondence, but rather the class could be expressed by the presence or absence of multiple features.
Even within the same class, the features may differ in terms of letter thickness, size, slant, etc.
Therefore, the feature of class $a$ could not be completely removed from $\bm{p}^{\{a, b\}}_i$, causing the occurrence rate of $\bm{p}^{\{a, b\}}_i \land \neg \bm{p}^{\{a\}}_i$ to be larger.
In addition, it is possible that there are common features in classes $a$ and $b$, and that $\bm{p}^{\{a, b\}}_i \land \neg \bm{p}^{\{b\}}_i$ also remove common features between the classes: thus the occurrence rates became low.
Therefore, we also investigate the top occurrence rate using query $\bm{p}^{\{a, b\}}_i \land \stackrel{\bm{p}_i^{\{a\}}}{\lnot} \bm{p}^{\{b\}}_i$, in which the features that only $\bm{p}^{\{b\}}_i$ has in $\bm{p}^{\{a\}}_i,\ \bm{p}^{\{b\}}_i$ are removed from $\bm{p}^{\{a, b\}}_i$.
The operator $\stackrel{\bm{p}_i^{\{a\}}}{\lnot} \bm{p}^{\{b\}}_i$ is defined as follows:
\begin{equation}
    \stackrel{\bm{p}_i^{\{a\}}}{\lnot} \bm{p}^{\{b\}}_i := \lnot \bm{p}^{\{b\}}_i \lor (\bm{p}^{\{a\}}_i \land \bm{p}^{\{b\}}_i)
\end{equation}
%
Since the common feature $\bm{p}^{\{a\}}_i \land \bm{p}^{\{b\}}_i$ is not removed from $\bm{p}^{\{a, b\}}_i$ , the value of $\bm{p}^{\{a, b\}}_i \land \stackrel{\bm{p}_i^{\{a\}}}{\lnot} \bm{p}^{\{b\}}_i$ is closer to the reference value than $\bm{p}^{\{a, b\}}_i \land \neg \bm{p}^{\{b\}}_i$. 
%
%
\begin{figure*}[t]
    \centering
    \includegraphics[width=0.95\linewidth]{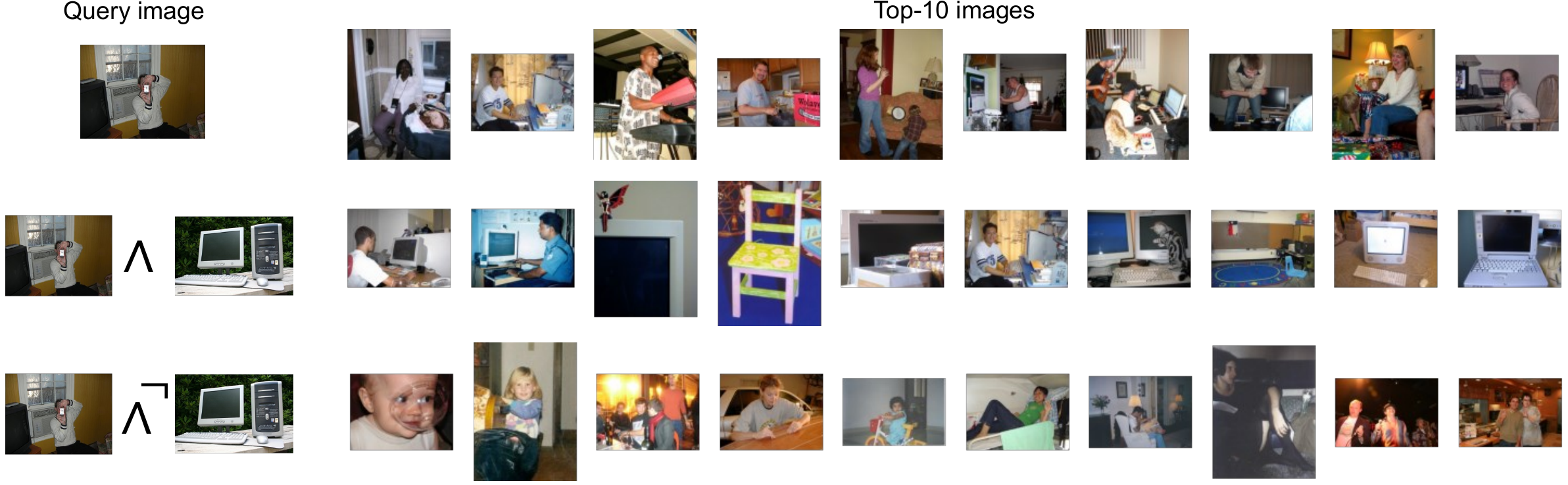}
    \caption{
        \textbf{Examples of the top-10 images of image retrieval.}
        The top row displays the images retrieved using a query for an image containing both 'tvmonitor' and 'person'. 
        The middle row presents results from an AND operation between images containing both 'tvmonitor' and 'person' and those containing only 'tvmonitor'. 
        This shows an increase in images containing only 'tvmonitor' compared to the top row. 
        The results in the bottom row are derived from an AND operation between the representation of images containing both 'tvmonitor' and 'person' and the negation of the representation of images containing only 'tvmonitor'. 
        This operation has resulted in the removal of the 'tvmonitor' feature, and an increase in images that contain only 'person'
        }
    \label{fig:and_or_examples}
\end{figure*}
\subsubsection{PascalVOC}
Futher, we evaluated the operability of the representation on PascalVOC, a multilabel dataset of natural images.
We defined $L$ as the set of classes with at least 200 images containing only one class from the trainval2007 and trainval2012 sets.
In addition, we defined sets of images in test2007 set that contain only class $a \in L$ or class $b \in L$ are defined as $X_{\{a\}}$ and $X_{\{b\}}$.
We also define a set $X_{\{a, b\}}$ of images in test2007 set that contains only class $a$ and class $b$.
We prepared the model pretrained by trainval2007 and trainval2012 sets of PascalVOC by 500 epochs.
A gallery set of 2400 images was created by sampling 200 images from the trainval2007 and trainval2012 sets for each class.
Using the classes, image sets, and models defined above, we compared the occurrence rates of class $a$ in similar experiments in Sec.~\ref{sec:image_retrieval}-\ref{sec:exp_mnist}

Fig.~\ref{fig:image_retrieval_rerult_rate_voc} shows the average occurrence rate.
Similar to the MNIST results, we can confirm that OR and AND are working properly by comparing $\bm{p}^{\{a, b\}}_i$ to $\bm{p}^{\{a\}}_i \lor \bm{p}^{\{b\}}_i$, $\bm{p}^{\{a\}}_i$ to $\bm{p}^{\{a, b\}}_i \land \bm{p}^{\{a\}}_i$, and $\bm{p}^{\{b\}}_i$ to $\bm{p}^{\{a, b\}}_i \land \bm{p}^{\{b\}}_i$.
The difference from the reference value appears to be larger than that of MNIST.
The reason may be that they are likely to have different features within the same class (e.g., gender and age for the ``person'' class, and breed for the ``dog'' class).

Fig.~\ref{fig:and_or_examples} shows some examples of image retrieval, including the result of AND and NOT operations between the representation of the image of ``tvmonitor'' and the representation of the image containing both ``tvmonitor'' and ``person''.
In the result of second row, many images of "tvmonitor" appear in the top-10 images, indicating that the representation possesses features related to "tvmonitor" owing to the AND operation.
On the other hand, in the third row, the AND, NOT operation removes the features related to "tvmonitor", and many "person" images appear in the top-10 images.
\section{LIMITATION}
The computational cost of our proposed representation synthesis by logic operations is higher than that by mean and maximum operations.
When a representation consists of $N$ $M$-class categorical distributions, the computational cost required for synthesis is $O(NM)$ for the mean and maximum operations and $O(NM^2)$ for the logic operations.
We note that our method increases the computational cost, especially in the synthesis of the representations, but this increase does not affect the gradient computation during the learning of the encoder part.
Although there is an increase in computational cost, this increase is minimal.

This study focused on only image representation.
However, if the operability discussed in this study can be extended to multimodal representation learning using images and language in the future, the controllability of representation can be improved.
\section{CONCLUSION}
In this study, we proposed a self-supervised representation learning method that can perform logic operations by describing the degree of possession of each feature in an image using many-valued logic.
The results showed that the performance of our method was competitive with those of other SSLs that use representation synthesis in classification tasks.
Further, we experimentally verified that the possession degree of each feature in the image was properly learned.
Moreover, we showed that the proposed method can learn representations with high logic-operability using image retrieval tasks.

\bibliographystyle{IEEEbib}
\bibliography{egbib.bib}

\vfill\pagebreak
\end{document}